\documentclass[runningheads]{llncs}
\usepackage{graphicx}
\usepackage{amsmath}
\usepackage{amssymb}
\usepackage{subcaption}
\usepackage{verbatim}
\usepackage{cite}
\usepackage{color}

\usepackage{adjustbox}
\begin{document}
\title{Skin Lesion Synthesis with\\Generative Adversarial Networks}
\author{Alceu Bissoto$^1$ \and F\'abio Perez$^2$ \and Eduardo Valle$^2$ \and Sandra Avila$^1$\vspace{-0.1cm}}
\institute{$^1$RECOD Lab, IC, University of Campinas (Unicamp), Brazil\\$^2$RECOD Lab, DCA, FEEC, University of Campinas (Unicamp), Brazil\vspace{-0.2cm}}
\authorrunning{A. Bissoto et al.}

\maketitle              
\begin{abstract}

Skin cancer is by far the most common type of cancer. 
Early detection is the key to increase the chances for successful treatment significantly.
Currently, Deep Neural Networks are the state-of-the-art results on automated skin cancer classification. 
To push the results further, we need to address the lack of annotated data, which is expensive and require much effort from specialists. 
To bypass this problem, we propose using Generative Adversarial Networks for generating realistic synthetic skin lesion images. 
To the best of our knowledge, our results are the first to show visually-appealing synthetic images that comprise clinically-meaningful information.

\keywords{Skin Cancer \and Generative Models \and Deep Learning\vspace{-0.2cm}}
\end{abstract}

\begin{figure}[t]
\centering

\begin{subfigure}[b]{\linewidth}
\centering
     \includegraphics[trim= 6cm 1cm 6cm 1cm,clip, width=0.23\textwidth]{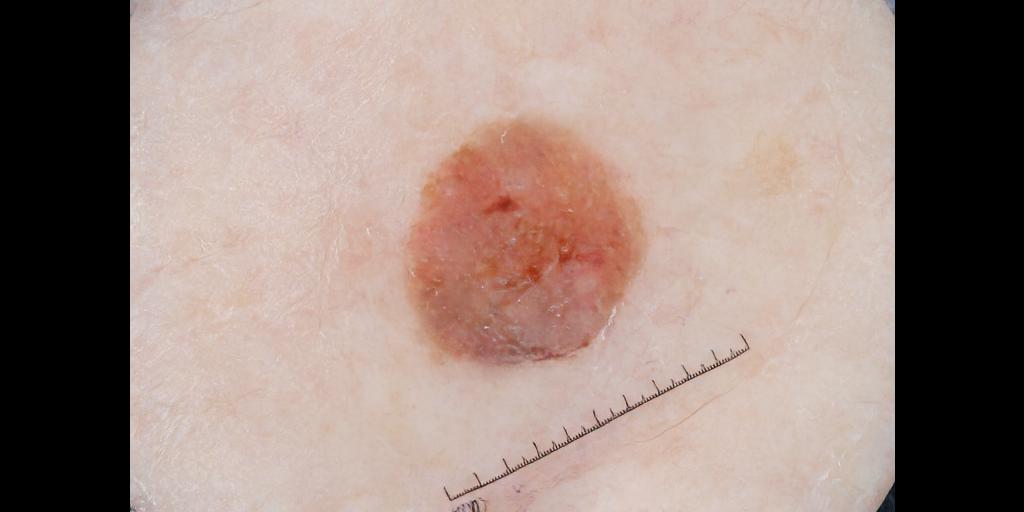}
    \includegraphics[trim= 6cm 1cm 6cm 1cm,clip, width=0.23\textwidth]{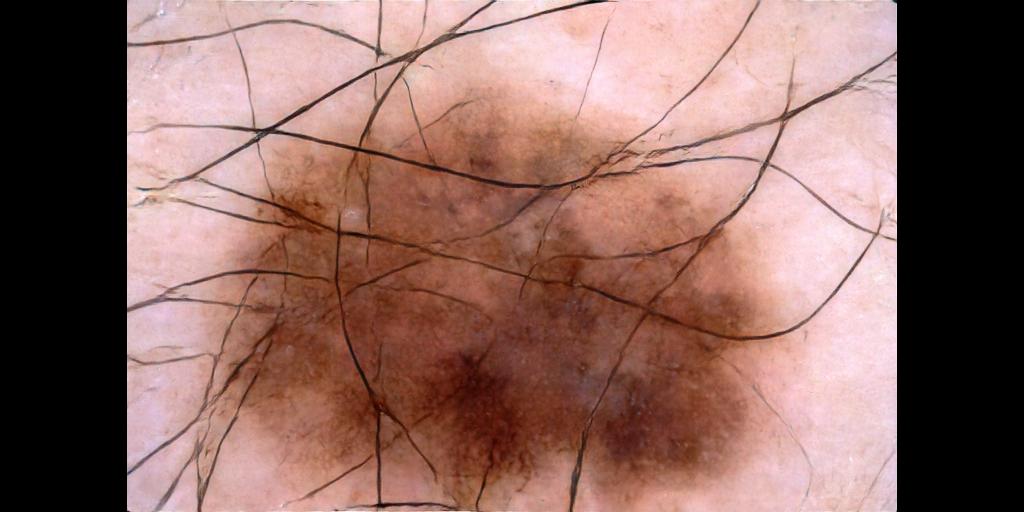}
    \includegraphics[trim= 6cm 1cm 6cm 1cm,clip, width=0.23\textwidth]{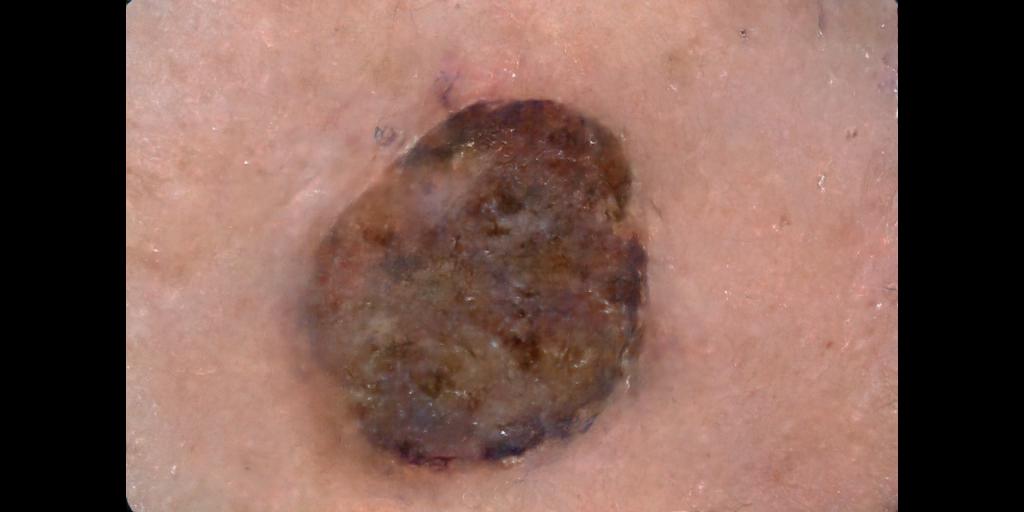}
    \includegraphics[trim= 6cm 1cm 6cm 1cm,clip, width=0.23\textwidth]{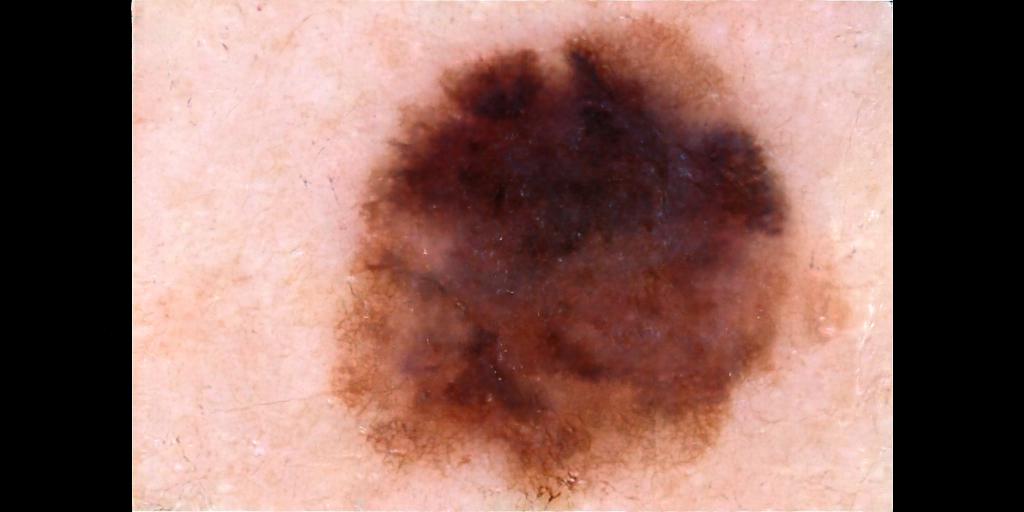}   
\end{subfigure}
\\

\vspace{-0.15cm}
\caption{Our approach successfully generates high-definition, visually-appealing, clinically-meaningful synthetic skin lesion images. All samples are synthetic. Details can be found in Sec.~\ref{method}.\vspace{-0.4cm}}\label{fig:samples}
\end{figure}

\section{Introduction}

Melanoma is the most dangerous form of skin cancer. It causes the most deaths, representing about 1\% of all skin cancers in the United States\footnote{http://www.cancer.net/cancer-types/melanoma/statistics}. The crucial point for treating melanoma is early detection. The estimated 5-year survival rate of diagnosed patients rises from 15\%, if detected in its latest stage, to over 97\%, if detected in its earliest stages
\cite{survival-rates-skin-cancer}. 

Automated classification of skin lesions using images is a challenging task owing to the fine-grained variability in the appearance of skin lesions. Since the adoption of Deep Neural Networks (DNNs), the state of the art improved rapidly for skin cancer classification~\cite{fornaciali2016towards,esteva2017,valle2018data,perez2018data}. 
To push forward, we need to address the lack of annotated data, which is expensive and require much effort from specialists. To bypass this problem, we propose using Generative Adversarial Networks (GANs)~\cite{goodfellow2014} for generating realistic synthetic skin lesion images.

GANs aim to model the real image distribution by forcing the synthesized samples to be indistinguishable from real images. Built upon these generative models, many methods were proposed to generate synthetic images based on GANs~\cite{radford2015dcgan,salimans2016,isola2017image}. A drawback of GANs is the resolution of the synthetic images~\cite{wang2017high}.
The vast majority of works is evaluated on low-resolution datasets such as CIFAR ($32\times32$) and MNIST ($28\times28$). However, for skin cancer classification, the images must have a higher level of detail (high resolution) to be able to display malignancy markers that differ a benign from a malignant skin~lesion.

Very few works have shown promising results for high-resolution image generation.
For example, Karras~et~al.'s~\cite{karras2017progressive} progressive training procedure generate celebrity faces up to $1024\times1024$ pixels.
They start by feeding the network with low-resolution samples. Progressively, the network receives increasingly higher resolution training samples while amplifying the respective layers' influence to the output.  
In the same direction, Wang et al.~\cite{wang2017high} generate high-resolution images from semantic and instance maps. They propose to use multiple discriminators and generators that operate in different resolutions to evaluate fine-grained detail and global consistency of the synthetic samples.
We investigate both networks for skin lesion synthesis, comparing the achieved results.

In this work, we propose a GAN-based method for generating high-definition, visually-appealing, and clinically-meaningful synthetic skin lesion images. To the best of our knowledge, this work is the first that successfully generates realistic skin lesion images (for illustration, see Fig.~\ref{fig:samples}). To evaluate the relevance of synthetic images, we train a skin cancer classification network with synthetic and real images, reaching an improvement of 1 percentage point. Our full implementation is available at \url{https://github.com/alceubissoto/gan-skin-lesion}.\vspace{-0.3cm}

\section{Proposed Approach} \label{method}
\vspace{-0.1cm}

Our aim is to generate high-resolution synthetic images of skin lesions with fine-grained detail. 
To explicitly teach the network the malignancy markers while incorporating the specificities of a lesion border, we feed these information directly to the network as input. Instead of generating the image from noise (usual procedure with GANs), we synthesize from a semantic label map (an image where each pixel value represents the object class) and an instance map (an image where the pixels combine information from its object class and its instance).
Therefore, our problem of image synthesis specified to image-to-image translation. 

\subsection{GAN Architecture: The pix2pixHD Baseline}
\vspace{-0.05cm}

We employ Wang's et al.~\cite{wang2017high} pix2pixHD GAN, which improve the pix2pix network \cite{isola2017image} (a conditional image-to-image translation GAN) by using a coarse-to-fine generator, a multi-scale discriminator architecture, and a robust adversarial learning objective function. The proposed enhancements allowed the network to work with high-resolution samples.

\begin{figure}[t]
\centering
\includegraphics[width=\textwidth]{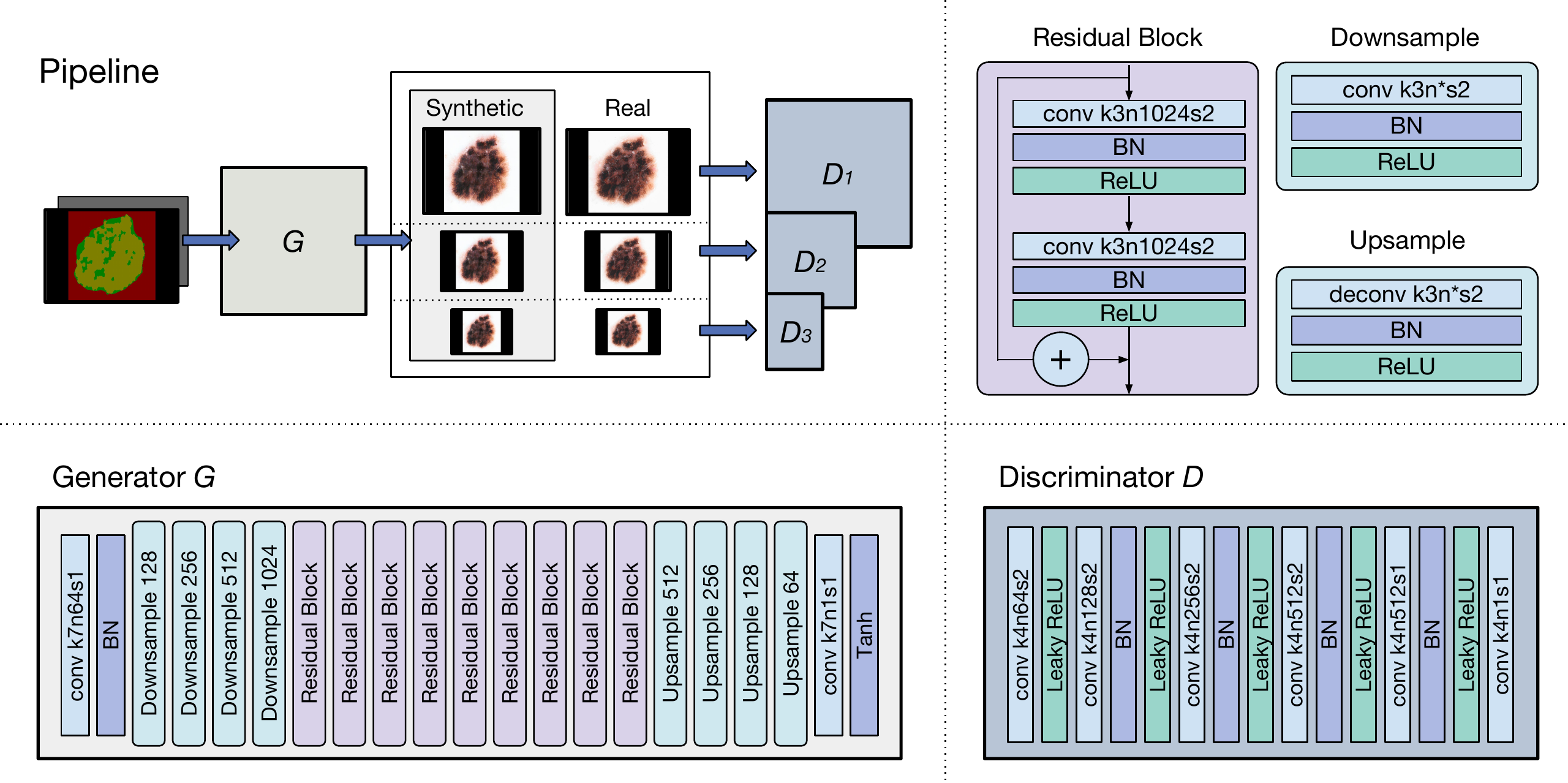}
\vspace{-0.5cm}
\caption{Summary of the GAN architecture. In the bottom-left, we show the pipeline. We detail both discriminator and generator, and the blocks that compose them. We show the parameters for each convolutional layer: $k$ is the kernel size; $n$ is the number of channels; and $s$ is the stride. The number that follows both Downsample and Upsample blocks are the numbers of channels.\vspace{-0.3cm}}\label{architecture}
\end{figure}

For generating $1024\times512$ resolution images, we only take advantage of the Global generator from pix2pixHD. This generator's output resolution fits with the minimum common size of our dataset images.
It is composed of a set of convolutional layers, followed by a set of residual blocks \cite{he2016deep} and a set of deconvolutional layers. 

To handle global and finer details, we employ three discriminators as Wang et al.~\cite{wang2017high}. 
Each of the three discriminators receives the same input in different resolutions. This way, for the second and third discriminator, the synthetic and real images are downsampled by 2 and 4 times respectively. Fig. \ref{architecture} summarizes the architecture of the GAN network. 

The loss function incorporates the feature matching loss \cite{salimans2016} to stabilize the training. It compares features of real and synthetic images from different layers of all discriminators. The generator learns to create samples that match these statistics of the real images at multiple scales.
This way, the loss function is a combination of the conditional GAN loss, and feature matching loss.\vspace{-0.1cm}

\subsection{Modeling Skin Lesion Knowledge}

Modeling meaningful skin lesion knowledge is the crucial condition for synthesizing high-quality and high-resolution skin lesions images. In the following, we show how we model the skin lesion scenario into semantic and instance maps for image-to-image translation.

\textbf{Semantic map}~\cite{lin2014microsoft} is an image where every pixel has the value of its object class and is commonly seen as a result of pixel-wise segmentation tasks.

To compose our semantic map, we propose using masks that show the presence of five malignancy markers and the same lesions' segmentation masks.  
The skin without lesion, the lesion without markers, and each malignancy marker are assigned a different label. To keep the aspect ratio of the lesions, while keeping the size of the input constant as the same of the original implementation by Wang et al.~\cite{wang2017high}, we assign another label to the borders, which do not constitute the skin image.

\textbf{Instance map}~\cite{lin2014microsoft} is an image where the pixels combine information from its object class and its instance. Every instance of the same class receives a different pixel value. When dealing with cars, people, and trees, this information is straightforward, but to structures within skin lesions, it is subjective. 

\begin{figure}
\centering
\begin{subfigure}[b]{0.29\textwidth}
\includegraphics[width=\textwidth]{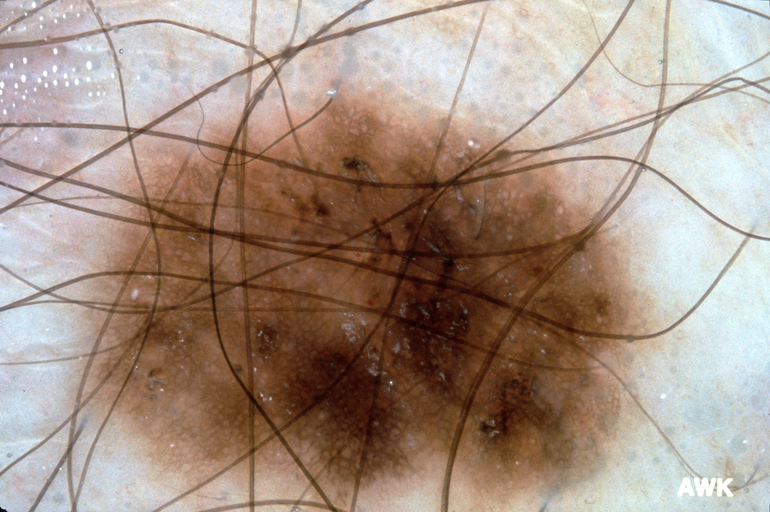}
\caption{Real image}
\end{subfigure}
\begin{subfigure}[b]{0.29\textwidth}
\includegraphics[width=\textwidth]{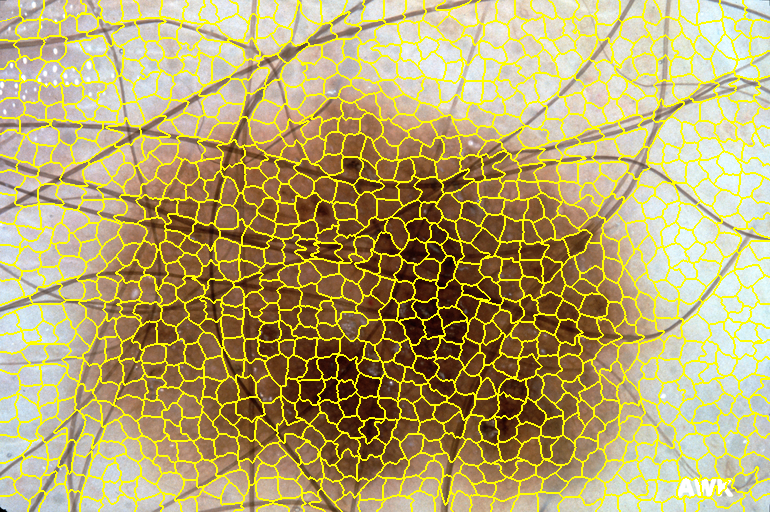}
\caption{Superpixels}
\end{subfigure}
\begin{subfigure}[b]{0.29\textwidth}
\includegraphics[width=\textwidth]{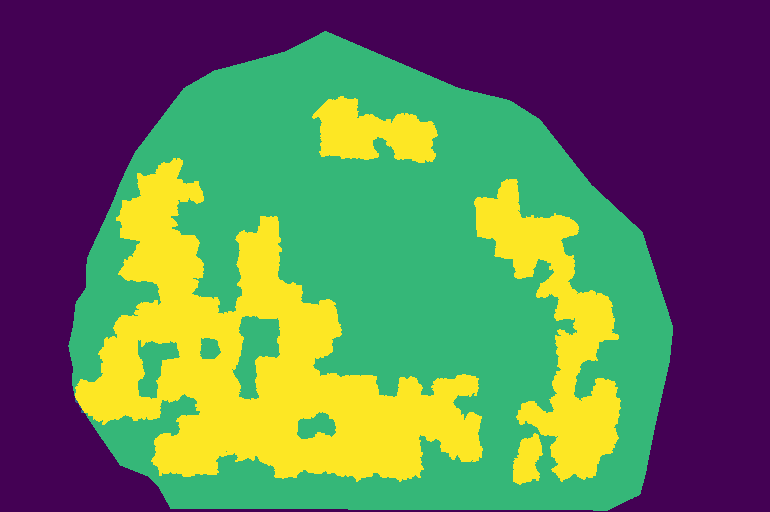}
\caption{Semantic label map}
\end{subfigure}
\vspace{-0.125cm}
\caption{A lesion's semantic map, and its superpixels representing its instance map. Note how superpixels change its shape next to hairs and capture information of the lesion borders, and interiors.\vspace{-1cm}}\label{masks}
\end{figure}

To compose our instance maps, we take advantage of superpixels \cite{achanta2010slic}. \textit{Superpixels} group similar pixels creating visually meaningful instances.
They are used in the process of annotation of the malignancy markers masks. First, the SLIC algorithm \cite{achanta2010slic} is applied to the lesion image to create the superpixels. Then, specialists annotate each of the superpixels with the presence or absence of five malignancy markers. Therefore, superpixels are the perfect candidate to differentiate individuals within each class, since they are already in the annotation process as the minimum unit of a class. In Fig. \ref{masks} we show a lesion's semantic map, and its superpixels representing its instance map.

Next, we conduct experiments to analyze our synthetic images and compare the different approaches introduced to generate them.\vspace{-0.25cm}

\section{Experiments} \label{experiments}
\vspace{-0.1cm}

In this section, we evaluate GAN-based approaches for generating synthetic skin lesion images: 1) DCGAN~\cite{radford2015dcgan}, 2) our conditional version of PGAN \cite{karras2017progressive}, and 3) our versions of pix2pixHD \cite{wang2017high} using only semantic map, and 4)~using semantic and instance maps. We choose DCGAN to represent low-resolution GANs because of its traditional architecture. Results for other low-resolution GANs do not show much of an improvement.\vspace{-0.25cm}
\subsection{Datasets}

For training and testing pix2pixHD, we need specific masks that show the presence or absence of clinically-meaningful skin lesion patterns (including pigment network, negative network, streaks, mlia-like cysts, and globules). These masks are available from the training dataset of task~2 (2,594 images) of 2018 ISIC Challenge\footnote{\url{https://challenge2018.isic-archive.com}}. The same lesions' segmentation masks that are used to compose both semantic and instance maps were obtained from task~1 of 2018 ISIC Challenge. 
We split the data into train (2,346 images) and test (248 images). 
The test is used for generating images using masks the network has never seen before. 

For training DCGAN and our version of PGAN, we use the following datasets:
ISIC 2017 Challenge with 2,000 dermoscopic images~\cite{codella2017skin},
ISIC Archive with 13,000 dermoscopic images,
Dermofit Image Library \cite{ballerini2013color} with 1,300 images, and
PH2 dataset \cite{mendoncca2013ph} with 200 dermoscopic image.

For training the classification network, we only use the 'train' set (2,346~images).
For testing, we use the Interactive Atlas of Dermoscopy~\cite{argenziano2002dermoscopy} with 900 dermoscopic images (270 melanomas).\vspace{-0.1cm}

\subsection{Experimental Setup}
For pix2pixHD, DCGAN (official PyTorch implementation) and PGAN (except for the modifications listed below), we keep the default parameters of each implementation.

We modified PGAN by concatenating the label (benign or melano\-ma) in every layer except the last on both discriminator and generator. For training, we start with $4\times4$ resolution, always fading-in to the next resolution after 60 epochs, from which 30 epochs are used for stabilization. To generate images of resolution $256\times256$, we trained for 330 epochs. We ran all experiments using the original Theano version.  

For skin lesion classification, we employ the network (Inception-v4~\cite{szegedy2016inceptionv4}) ranked first place for melanoma classification~\cite{recod2017} at the ISIC 2017 Challenge.
As Menegola et al.~\cite{recod2017}, we apply random vertical and horizontal flips, random rotations and color variations as data augmentation. Also we keep test augmentation with 50 replicas, but skip the meta-learning SVM.\vspace{-0.1cm}

\subsection{Qualitative Evaluation}
\begin{figure}[h!]
\centering
\begin{subfigure}[b]{\linewidth}
\centering
	\hspace{-0.1cm}\includegraphics[trim= 0 0.55cm 0 0.55cm, clip, width=0.19\textwidth]{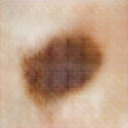}
	\includegraphics[trim= 0 1.05cm 0 1.22cm, clip, width=0.19\textwidth]{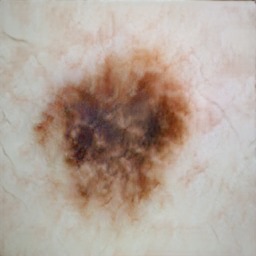} 	  \includegraphics[trim= 6cm 0 6cm 0,clip, width=0.19\textwidth]{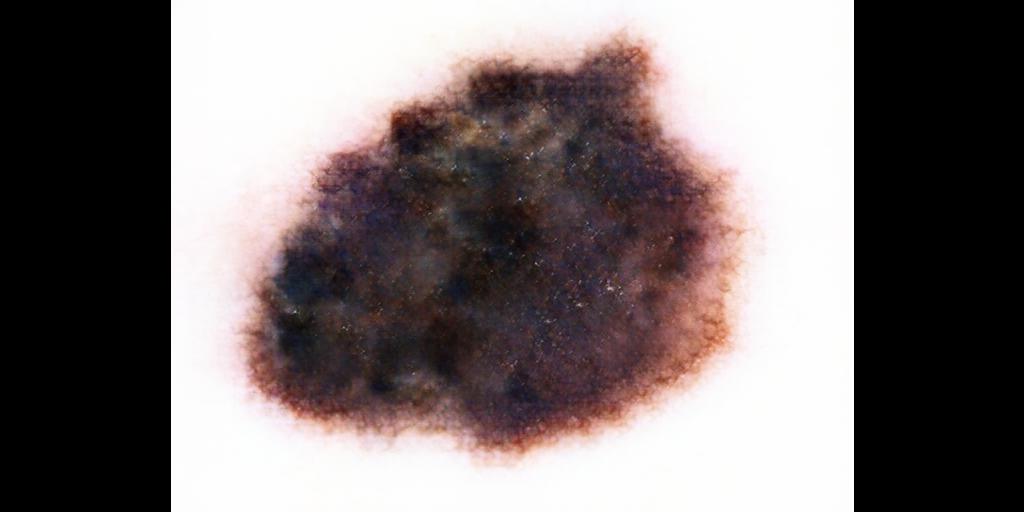}
	\includegraphics[trim= 6cm 0 6cm 0,clip, width=0.19\textwidth]{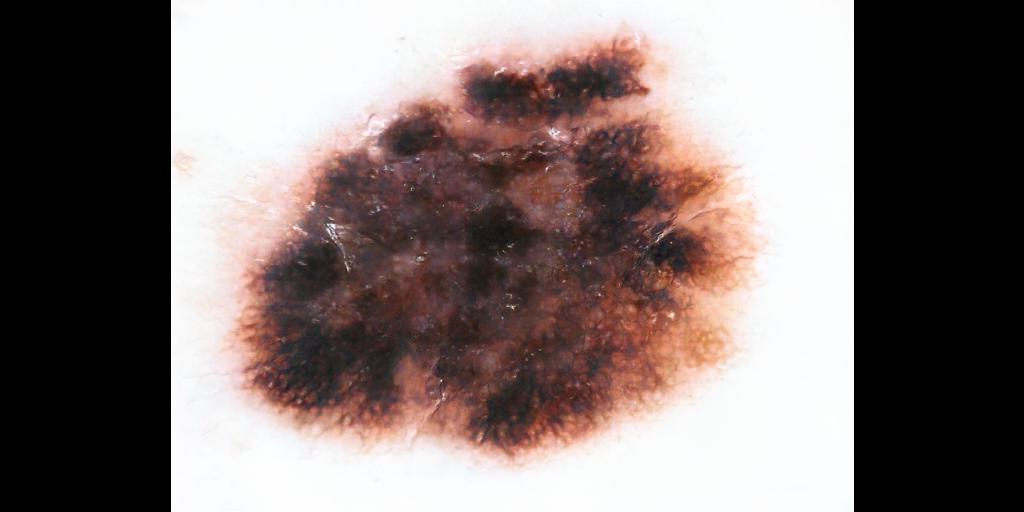}
	\includegraphics[width=0.19\textwidth]{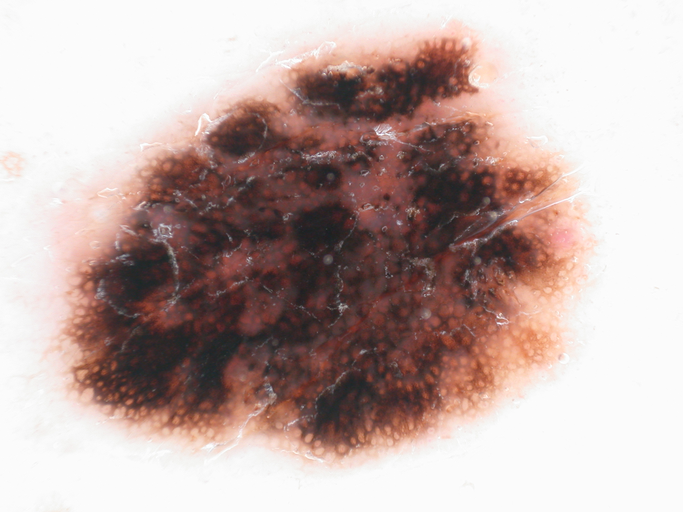}
\end{subfigure}
\\\vspace{1mm}
\begin{subfigure}[b]{0.19\textwidth}
	\includegraphics[trim= 0 1.5mm 0 0.45mm, clip, width=\textwidth]{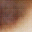}
    \caption{DCGAN}\label{gans-dcgan}
\end{subfigure}
\begin{subfigure}[b]{0.19\textwidth}
	\includegraphics[trim= 0 3mm 0 0.9mm, clip, width=\textwidth]{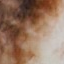}
    \caption{Ours}\label{gans-progressive}
\end{subfigure}
\begin{subfigure}[b]{0.19\textwidth}
    \includegraphics[trim= 0 6mm 0 1.8mm,clip, width=\textwidth]{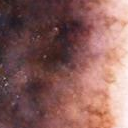}
    \caption{Ours}\label{gans-semantic}
\end{subfigure}
\begin{subfigure}[b]{0.19\textwidth}
	\includegraphics[trim= 0 6mm 0 1.8mm,clip, width=\textwidth]{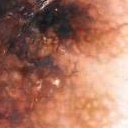}
    \caption{Ours}\label{gans-instance-semantic}
\end{subfigure}
\begin{subfigure}[b]{0.19\textwidth}
	\includegraphics[trim= 0 6mm 0 1.8mm,clip, width=\textwidth]{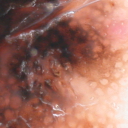}
    \caption{Real}
\end{subfigure}
\vspace{-0.1cm}
\caption{Results for different GAN-based approaches: (a) DCGAN~\cite{radford2015dcgan}, (b) Our version of PGAN, (c) Our version of pix2pixHD using only semantic map, (d) Our version of pix2pixHD using both semantic and instance map, (e)~Real image. In the first row, we present the full image while in the second we zoom-in to focus on the details. 
\vspace{-0.35cm}}\label{gans}
\end{figure}

In Fig. \ref{gans} we visually compare the samples generated by GAN-based approaches.  

DCGAN (Fig.~\ref{gans-dcgan}) is one of the most employed GAN architectures. We show that samples generated by DCGAN are far from the quality observed on our models. It lacks fine-grained detail, being inappropriate for generating high-resolution samples.

Despite the visual result for PGAN (Fig.~\ref{gans-progressive}) is better than any other work we know of, it lacks cohesion, positioning malignancy markers without proper criteria. We cannot pixel-wise compare the PGAN result with the real image. This synthetic image was generated from noise and had no connection with the sampled real image, except it was part of the GAN's training set. But, we can compare the sharpness, the presence of malignancy markers and their fine-grained details.

When we feed the network with semantic label maps (Fig.~\ref{gans-semantic}) that inform how to arrange the malignancy markers, the result improves remarkably. When combining both semantic and instance maps (Fig.~\ref{gans-instance-semantic}), we simplify the learning process, achieving the overall best visual result. The network learns patterns of the skin, and of the lesion itself.\vspace{-0.2cm}

\subsection{Quantitative Evaluation}
To evaluate the complete set of synthetic images, we train a skin classification network with real and synthetic training sets and compare the area under the ROC curve (AUC) when testing only with real images. We use three different synthetic images for this comparison: \textbf{Instance} are the samples generated using both semantic and instance maps with our version of pix2pixHD \cite{wang2017high}; \textbf{Semantic} are the samples generated using only semantic label maps; \textbf{PGAN} are the samples generated using our conditional version of PGAN \cite{karras2017progressive}. For statistical significance, we run each experiment 10 times.

For every individual set, we use 2,346 images, which is the size of our training set (containing semantic and instance maps) for pix2pixHD. For PGAN, there is not a limitation in the amount of samples we are able to generate, but we keep it the same maintaining the ratio between benign and malignant lesions. Our results are in Table~\ref{auc}. To verify statistical significance (comparing `Real + Instance + PGAN' with other results), we include the p-value of a paired samples t-test.
With a confidence of 95\%, all differences were significant (p-value $<$ 0.05).

\begin{table}[t]
\caption{Performance comparison of real and synthetic training sets for a skin cancer classification network. We train the network 10 times with each set.
The features present in the synthetic images are not only visually appealing but also contain meaningful information to correctly classify skin lesions.\vspace{-0.2cm}}\label{auc}
\begin{center}
\begin{tabular}{|l|c|c|c|}
\hline
~Training Data       & ~AUC (\%)  &  ~Training Data Size~ & ~p-value~ \\ 
\hline
~Real                & ~$83.4 \pm 0.9$~ & 2,346 & ~$2.5\times10^{-3}$~~\\ 
~Instance            & ~$82.0 \pm 0.7$~ & 2,346 & ~$2.8\times10^{-5}$~~\\
~Semantic            & ~$78.1 \pm 1.2$~ & 2,346 & ~$6.9\times10^{-8}$~~\\
~PGAN                & ~$73.3 \pm 1.5$~ & 2,346 & ~$2.3\times10^{-9}$~~\\
~Real+Instance       & ~$82.8 \pm 0.8$~ & 4,692 & ~$1.1\times10^{-4}$~~\\
~Real+Semantic       & ~$82.6 \pm 0.8$~ & 4,692 & ~$1.2\times10^{-4}$~~\\
~Real+PGAN           & ~$83.7 \pm 0.8$~ & 4,692 & ~$2.6\times10^{-2}$~~\\
~Real+2$\times$PGAN  & ~$83.6 \pm 1.0$~ & 7,038 & ~$2.0\times10^{-2}$~~\\
~Real+Instance+PGAN~ & ~$84.7 \pm 0.5$~ & 7,038 & ~--~~\\
\hline 
\end{tabular}
\end{center}
\vspace{-0.35cm}
\end{table}

The synthetic samples generated using instance maps are the best among the synthetics. The AUC follows the visual quality perceived. 

The results for synthetic images confirm they contain features that characterize a lesion as malignant or benign. Even more, the results suggest the synthetic images contain features that are beyond the boundaries of the real images, which improves the classification network by an average of 1.3 percentage~point and keeps the network more stable. 

To investigate the influence of the instance images over the achieved AUC for `Real + Instance + PGAN', we replace the instance images with new PGAN samples (`Real + 2$\times$PGAN'). Although both training sets have the same size, the result did not show improvements over its smaller version `Real + PGAN'. Hence, the improvement over the AUC achieved suggests it is related with the variations the `Instance' images carry, and not (only) by the size of the train~dataset.\vspace{-0.1cm}

\section{Conclusion} \label{Conclusion}
\vspace{-0.1cm}

In this work, we propose GAN-based methods to generate realistic synthetic skin lesion images. 
We visually compare the results, showing high-resolution samples (up to $1024\times512$) that contain fine-grained details. Malignancy markers are present with coherent placement and sharpness which result in visually-appealing images. We employ a classification network to evaluate the specificities that characterize a malignant or benign lesion. The results show that the synthetic images carry this information, being appropriate for classification purposes. 

Our pix2pixHD-based solution, however, requires annotated data to generate images. To overcome this limitation, we are working on different approaches to generate diversified images employing pix2pixHD without additional data: combining different lesions' semantic and instance masks, distorting existing real masks for creating new ones, or even employing GANs for the easier task of generating masks. Despite the method used, taking advantage of synthetic images for classification is promising.

\section*{Acknowledgments}

We gratefully acknowledge NVIDIA for the donation of GPUs, Microsoft Azure for the GPU-powered cloud platform, and CCES/Unicamp (Center for Computational Engineering \& Sciences) for the GPUs used in this work. A. Bissoto is funded by CNPq.
E. Valle is partially funded by Google Research LATAM 2017, CNPq PQ-2 grant (311905/2017-0), and Universal grant (424958/2016-3). RECOD Lab. is partially supported by FAPESP, CNPq, and CAPES.

\bibliographystyle{splncs04}
\bibliography{refs}

\end{document}